\documentclass{article}

\PassOptionsToPackage{numbers,sort&compress}{natbib}

\usepackage[main,final]{neurips_2026}

\usepackage[utf8]{inputenc} 
\usepackage[T1]{fontenc}    
\usepackage{courier}        
\usepackage{hyperref}       
\usepackage{url}            
\usepackage{booktabs}       
\usepackage{amsfonts}       
\usepackage{nicefrac}       
\usepackage{microtype}      
\usepackage{xcolor}         

\usepackage{multirow}
\usepackage{amsmath}
\usepackage{graphicx}
\usepackage{wrapfig}
\title{Pinpoint: Grounded Worldwide Image Geolocation via Cross-Source Retrieval and Reranking}

\author{%
  Nika Chuzhoy$^{1,2}$\thanks{Equal contribution.} \quad
  Brian Hu$^{1,2}$\footnotemark[1] \quad
  Amit A. Arora$^{1}$ \quad
  Jae Ro$^{1}$ \quad
  Sarthak S.~Sahu$^{1}$ \\
  $^{1}$Virtualitics \quad $^{2}$California Institute of Technology
}

\begin{document}

\maketitle

\begin{abstract}
  Image geolocation aims to estimate where a photograph was taken from its visual content. At worldwide scale, this remains challenging  because visual evidence is often ambiguous, diverse, and unevenly distributed. Prior work has typically treated geolocation of ordinary internet photos and street-view imagery as separate tasks, despite their complementary strengths: internet photos better match the appearance distribution of user-captured queries, while street-view imagery provides denser, geographically grounded coverage. We present Pinpoint, a retrieve-and-rerank architecture that combines both sources in a coarse-to-fine pipeline. A contrastive image--GPS embedder is trained on both user-uploaded Flickr photos and street-view imagery, learning a shared image--GPS embedding space that is used to retrieve candidate locations. An attention-based reranker then rescores retrieved candidates by combining candidate-level visual and GPS features with cross-source evidence from nearby locations to ground the prediction. Unlike recent prior work, Pinpoint does not rely on multimodal large-language models, making inference faster and more reproducible. Pinpoint achieves state-of-the-art results across all metrics on standard benchmarks for internet photos (IM2GPS3k and YFCC4k) and street-view imagery (OSV-5M).
\end{abstract}

\section{Introduction}
Worldwide image geolocation is the task of determining exactly where on Earth a photograph was taken. Reliable geolocation is valuable for a variety of applications such as navigation, disaster response, and open-source intelligence. However, worldwide geolocation is inherently challenging: the search space spans the entire globe, and many images lack distinctive landmarks or come from regions with sparse reference coverage. Recent state-of-the-art approaches rely on the strong world knowledge of multimodal large-language models (MLLMs) to resolve this ambiguity \cite{georanker,g3,img2loc,globe}. This strategy is powerful, but it imposes costs in latency and reproducibility, particularly for proprietary, externally hosted MLLMs. Further, it makes geolocation more dependent on latent knowledge acquired during internet-scale training, where benchmark contamination becomes a concern. We demonstrate with Pinpoint that MLLMs are not required for state-of-the-art performance.

Prior work has often separated the geolocation of ordinary internet photos and street-view imagery into distinct tasks, with each task served by models trained exclusively on data from its own regime. Internet-photo geolocation models target unconstrained images from sources such as Flickr, Wikimedia, or landmark collections \cite{georanker,pigeon,g3,geoclip}, where viewpoints and subjects vary widely. Street-view geolocation models instead focus on road-level panoramas or navigation datasets, where signs, road markings, infrastructure, and driving context provide more standardized cues \cite{pigeon, hierloc, aroundtheworld, osv5m}. This split has shaped both training data and evaluation: systems learn from a single source and are evaluated on benchmarks drawn from that same source, even though the two regimes contain complementary information. Ordinary internet photos expose models to diverse objects and scenes, while street-view imagery provides denser, more consistent geographic coverage.

We present Pinpoint, a retrieve-and-rerank architecture that implements cross-source training. Following prior work, we train Pinpoint separately for each evaluation setting---one instance targeting the internet-photo benchmarks and another targeting the street-view benchmark---but, unlike prior work, both instances draw on both data sources during training and exploit their complementary strengths. During the retrieval stage, Pinpoint uses an image tower based on a SigLIP~2~\cite{siglip2} backbone to embed Flickr photos and OSV-5M street-view images~\cite{osv5m} into a shared space. The Flickr images contribute diversity to the embedding space, while the street-view images contribute regularity. Street-view images are geographically denser and more homogeneous in appearance with respect to location, while Flickr images are more representative of in-the-wild photography.

During the reranking stage, Pinpoint uses an attention-based reranker trained specifically for candidate selection. For each retrieved top-$k$ location candidate, the reranker jointly attends to the query image representation, candidate embeddings, and a support token that summarizes nearby visual evidence around each candidate anchor location. Notably, neither the retrieval nor reranking stage uses MLLMs to generate or evaluate candidate locations. As a result, our model is readily reproducible, low-latency, and can be deployed privately.

We evaluate Pinpoint on the standard internet-photo benchmarks IM2GPS3k and YFCC4k, achieving state-of-the-art performance at every distance threshold on each, and we further set a new state of the art on the OSV-5M test set for street-view geolocation.

\section{Related Work}

\textbf{Worldwide Image Geolocation.} Many prior works have tackled the problem of global image geolocation, in three main ways. 
\textit{Classification-based methods} \cite{pigeon, planet, cplanet, translocator, geodecoder} treat geolocation as discrete prediction by partitioning the Earth into discrete regions or geocells and predicting the cell containing the query image. 
\textit{Retrieval-based methods} cast geolocation as a search over geotagged reference data \cite{geoclip, georanker,aerial,im2gps}. Early systems used hand-crafted features and nearest-neighbor aggregation \cite{im2gps}, while later methods adopted learned visual embeddings, contrastive objectives, and image--GPS alignment \cite{geoclip, g3, georanker}. While avoiding fixed geocells, retrieval methods can confuse visually similar but geographically distant places. Pinpoint retains retrieval's scalability to diverse data while addressing this visual ambiguity by introducing a reranking model that learns to finely distinguish between similar photos. \textit{MLLM-based methods} augment retrieval with generative vision-language models that leverage broad visual and world knowledge from internet-scale pretraining \cite{g3,georanker,img2loc}. These systems prompt MLLMs with retrieved geographic, textual, or visual evidence to generate or refine location predictions \cite{g3,georanker,img2loc}. As a result, MLLM-based approaches are often slower and non-deterministic at inference, and their reliance on proprietary models can make data contamination and reproducibility difficult to assess. In contrast, Pinpoint achieves stronger performance while keeping predictions grounded in a fixed retrieval database, making inference faster and deterministic. 

\textbf{Internet Photos and Street-View Imagery.}
Worldwide image geolocation has often treated ordinary internet photos and street-view imagery as separate settings. Internet-photo geolocation systems such as PlaNet~\cite{planet}, GeoCLIP~\cite{geoclip}, and PIGEOTTO~\cite{pigeon} target unconstrained web images from sources such as Flickr, Wikimedia, or landmark collections. Street-view geolocation systems instead exploit standardized road-level imagery \cite{osv5m, hierloc,aroundtheworld,semivariogramgeoclip}, where signs, lane markings, vegetation, architecture, and infrastructure provide dense geographic cues. 
Pinpoint combines these regimes: MP16-Pro, a dataset of images from Flickr \cite{g3}, supplies the inference-time retrieval index, while OSV-5M supplies dense auxiliary supervision during contrastive training.
We show that using street-view coverage improves geolocation of ordinary internet photos.

\textbf{Reranking for Retrieval.} Retrieve-and-rerank pipelines are
well-established in information retrieval, where a first-stage
retriever produces candidates that a slower, more expressive model
rescores \cite{bert,rerankingtransformers}. However, reranking is far from ubiquitous in geolocation models. The most closely related geolocation system to
our reranking approach is GeoRanker, which fine-tunes an MLLM with a learned distance-ranking objective to
rescore retrieved candidates \cite{georanker}.
Pinpoint implements reranking without an MLLM, using a lightweight attention-based reranker trained for geolocation.

\textbf{Visual Place Recognition.} Visual place recognition (VPR) is a related field focused on recognizing the same physical place despite changes in viewpoint and appearance. Methods such as NetVLAD~\cite{netvlad}, MixVPR~\cite{mixvpr}, and SALAD~\cite{salad} learn to aggregate visual features into global image descriptors, while R2Former~\cite{r2former} learns retrieval and reranking using local feature correspondences. VPR benchmarks such as Pitts250k~\cite{pitts250k} evaluate same-place retrieval at a 25 m threshold, whereas worldwide image geolocation evaluates geographic accuracy from 1 km to 2,500 km. At these scales, nearby images often do not depict overlapping views, particularly for heterogeneous internet photos which show both indoor and outdoor scenes. Pinpoint uses pooled SigLIP 2 features aligned with continuous GPS coordinates and reranks candidates using visual, geographic, and nearby cross-source evidence. Its image--GPS objective aligns each image with its own coordinate, producing GPS-aligned embeddings for image--image retrieval. This allows candidate selection to draw on geographic compatibility even when local visual correspondences are unavailable.

\section{Method}
\label{sec:method}
\begin{figure*}[t]
    \centering
    \includegraphics[width=\textwidth]{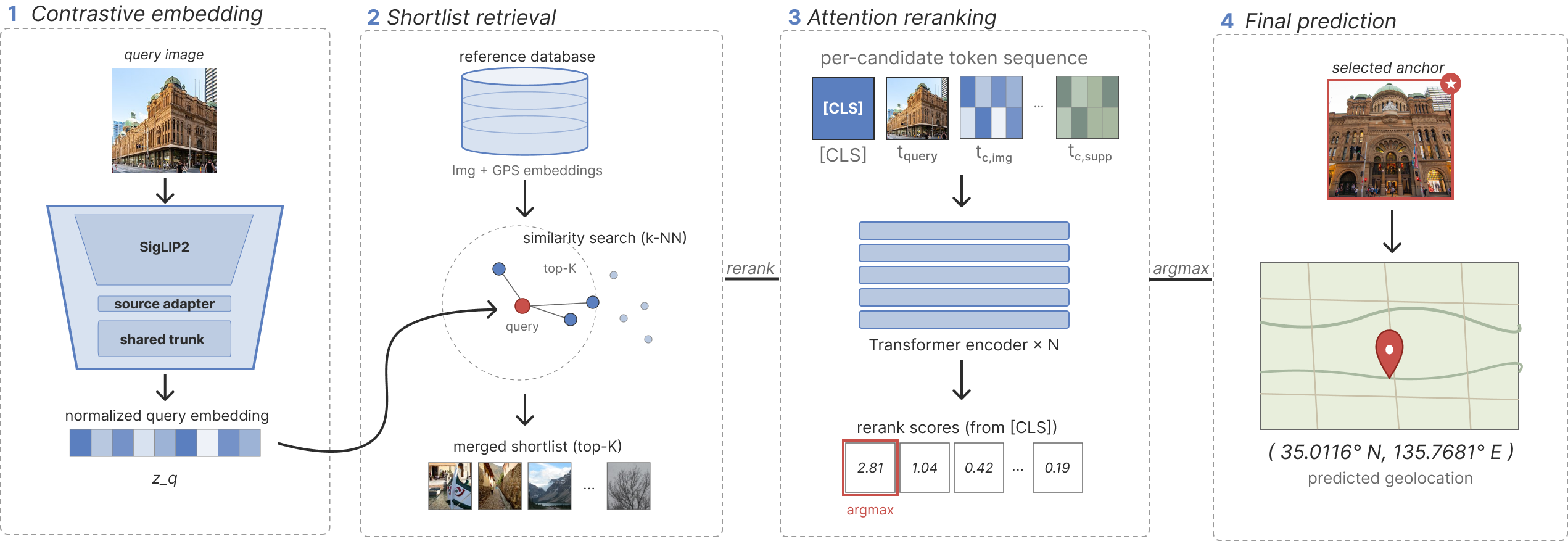}
    \caption{Overview of the Pinpoint inference pipeline. The query image is encoded by a frozen contrastive image tower and used to retrieve a top-$k$ shortlist from the retrieval index, which is then rescored by an attention-based reranker that selects the final predicted geolocation.}
    \label{fig:pipeline}
\end{figure*}

Pinpoint follows a two-stage retrieve-and-rerank pipeline, illustrated in Figure~\ref{fig:pipeline}. First, a contrastive image--GPS embedder maps images and GPS coordinates into a shared embedding space. Given a query image, this embedder retrieves a shortlist of candidate locations from a fixed retrieval index based on cosine similarity. During retrieval training, the model embeds images from both the street-view and internet-photo sources, learning to coarsely distinguish between a large number of diverse locations across the globe. Second, an attention-based reranker rescores each candidate by applying self-attention over a short sequence containing the query image embedding, the candidate's image embedding, SigLIP embedding, and GPS embedding along with a support token summarizing neighboring reference images from the complementary image source. This reranker model learns to finely distinguish between similar photos taken closer together. The final prediction is the GPS coordinate of the highest-scoring retrieved candidate.

\subsection{Contrastive Embedder}
\label{sec:embedder}
\begin{figure}[t]
    \centering
    \includegraphics[width=\columnwidth]{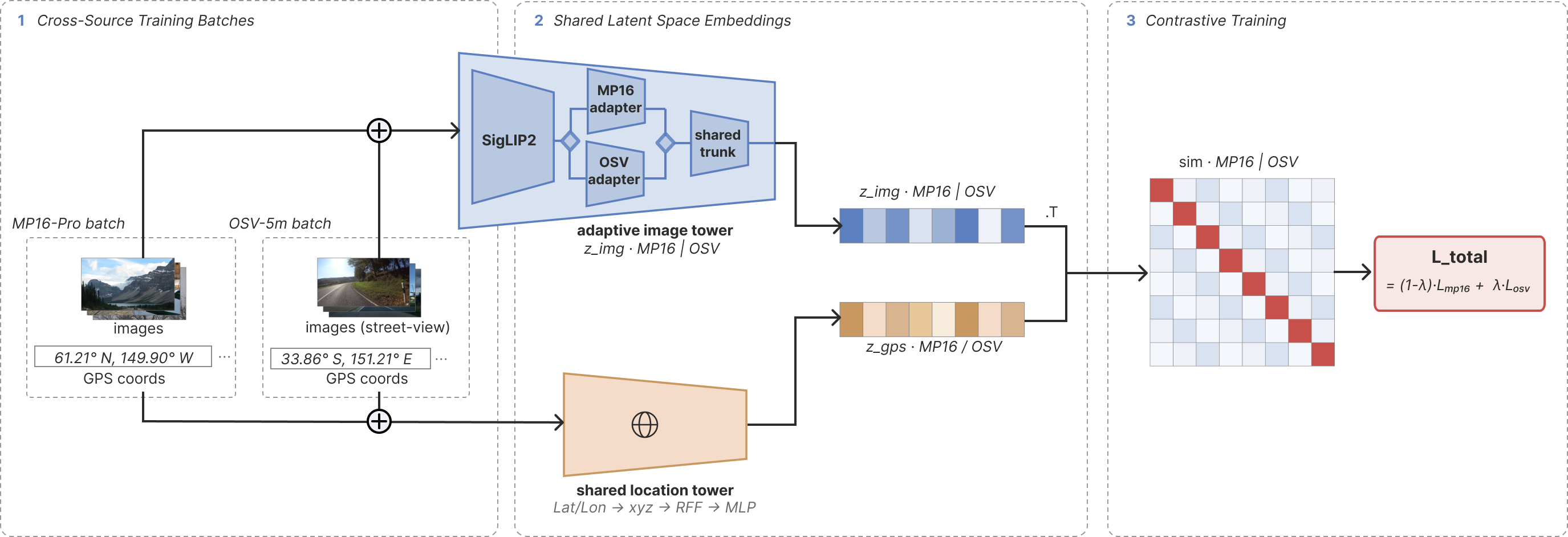}
    \caption{Cross-source image--GPS embedder. Image and GPS inputs are encoded by separate towers into a shared latent space and trained with a SigLIP contrastive objective over image--GPS pairs. The image tower consists of a frozen SigLIP backbone, source-specific input adapters, and a shared trunk. Total loss is a weighted sum of the source-specific contrastive losses.}
    \label{fig:embedder}
\end{figure}

The contrastive image--GPS embedder maps images and GPS coordinates into a shared $d$-dimensional latent space (Figure~\ref{fig:embedder}). Let $(x_{\text{img}}, x_{\text{gps}})$ denote an image and its GPS coordinate. The image and location towers produce $\ell_2$-normalized embeddings $z_{\text{img}}, z_{\text{gps}} \in \mathbb{R}^d$, trained so that each image's embedding has high cosine similarity to its true location embedding and low similarity to other location embeddings in the batch.

\paragraph{Location tower.} GPS coordinates given as (latitude, longitude) pairs are converted to 3D unit-sphere coordinates $(x, y, z) = (\cos\phi\cos\theta,\, \cos\phi\sin\theta,\, \sin\phi)$ to remove discontinuities at the poles and international date line, where $\phi$ is latitude and $\theta$ is longitude. The unit-vector representation is then passed through a multi-scale learned Fourier feature bank with $S$ scales. Each scale $s$ produces features $\bigl[\cos(2\pi W_s u + b_s),\, \sin(2\pi W_s u + b_s)\bigr]$ where $u$ is the unit vector and $W_s, b_s$ are learned parameters initialized from Random Fourier Features \cite{rff_orig}. The intuition is that low-frequency blocks encode global geographic structure while high-frequency blocks encode local variation.

For each scale, the unit-vector input is concatenated with that scale's Fourier features and passed through a residual MLP sub-tower. The per-scale outputs are combined with a softmax gate $g(u) \in \Delta^{S-1}$ whose weights are conditioned on the unit-vector position, allowing different regions of the sphere to weight scales differently:
\begin{equation}
h(u) \;=\; \sum_{s=1}^{S} g_s(u) \cdot \mathrm{Tower}_s\bigl(u, \phi_s(u)\bigr),
\end{equation}
where $\phi_s$ is the $s$-th Fourier feature block. A final linear projection produces $z_{\text{gps}} \in \mathbb{R}^d$.

\paragraph{Image tower.} We use SigLIP~2 (g-opt) as a frozen visual backbone  \cite{siglip2} and train only a projection tower on top of its cached image features. Because MP16-Pro and OSV-5M have different visual distributions, the projection tower uses source-specific input adapters followed by a shared trunk. Concretely, each training source $s \in \{\text{mp16}, \text{osv5m}\}$ is associated with its own input adapter $A_s : \mathbb{R}^{D_{\text{siglip}}} \to \mathbb{R}^{H_{\text{img}}}$, implemented as a small MLP. Both adapters feed into the same trunk and output projection. The adapters capture source-specific shifts between Flickr photos and street-view imagery, while the shared trunk learns a single geographically grounded representation common to both domains. At training time, each batch is source-tagged and routed through the appropriate adapter; the SigLIP image--GPS loss is computed separately for each source subset and then combined as a weighted sum. At inference time, the retrieval index contains task-specific images only (MP16-Pro or OSV-5M).

Intuitively, the OSV-5M images in the training dataset help regularize the shared embedding space to be more continuous and dense, while the MP16-Pro images help with diversity. Street-view images are all taken outside and are more homogeneous with respect to location; street-view images that are close geographically tend to look similar and have similar embeddings. Meanwhile, the distribution of internet photos is irregular with respect to location: a photo taken inside versus outside the same building can have very different embeddings. 

\paragraph{Training objective.} We train both towers jointly with the standard SigLIP sigmoid loss \cite{siglip}, which pulls matched image--GPS pairs together and pushes mismatched pairs apart. For a batch of $B$ matched image--GPS pairs, we form pairwise logits $\ell_{ij} = \alpha \, z_{\text{img},i}^{\top} z_{\text{gps},j} + \beta$ from image--location similarity, where $\alpha$ and $\beta$ are learnable scale and bias parameters, and set $y_{ij}=+1$ for matched pairs ($i=j$) and $y_{ij}=-1$ otherwise. The loss is
\begin{equation}
\mathcal{L}_{\text{SigLIP}} \;=\; \frac{1}{B} \sum_{i=1}^{B} \sum_{j=1}^{B} \log\!\bigl(1 + \exp(-y_{ij}\,\ell_{ij})\bigr).
\label{eq:siglip}
\end{equation}
The total embedder loss becomes the per-source weighted sum:
\begin{equation}
\mathcal{L}_{\text{total}} \;=\; (1-\lambda) \, \mathcal{L}_{\text{SigLIP}}\!\bigl(z_{\text{img}}^{(\text{mp16})}, z_{\text{gps}}^{(\text{mp16})}\bigr) \;+\; \lambda \, \mathcal{L}_{\text{SigLIP}}\!\bigl(z_{\text{img}}^{(\text{osv5m})}, z_{\text{gps}}^{(\text{osv5m})}\bigr),
\end{equation}
where $\lambda$ controls the contribution of the OSV-5M distribution relative to the MP16-Pro distribution. Appendix~\ref{app:retrieval_supervision} compares image--GPS and image--image supervision and their retrieval behavior.

\subsection{Attention Reranker}
\label{sec:reranker}
\begin{figure*}[t]
    \centering
    \includegraphics[width=\textwidth]{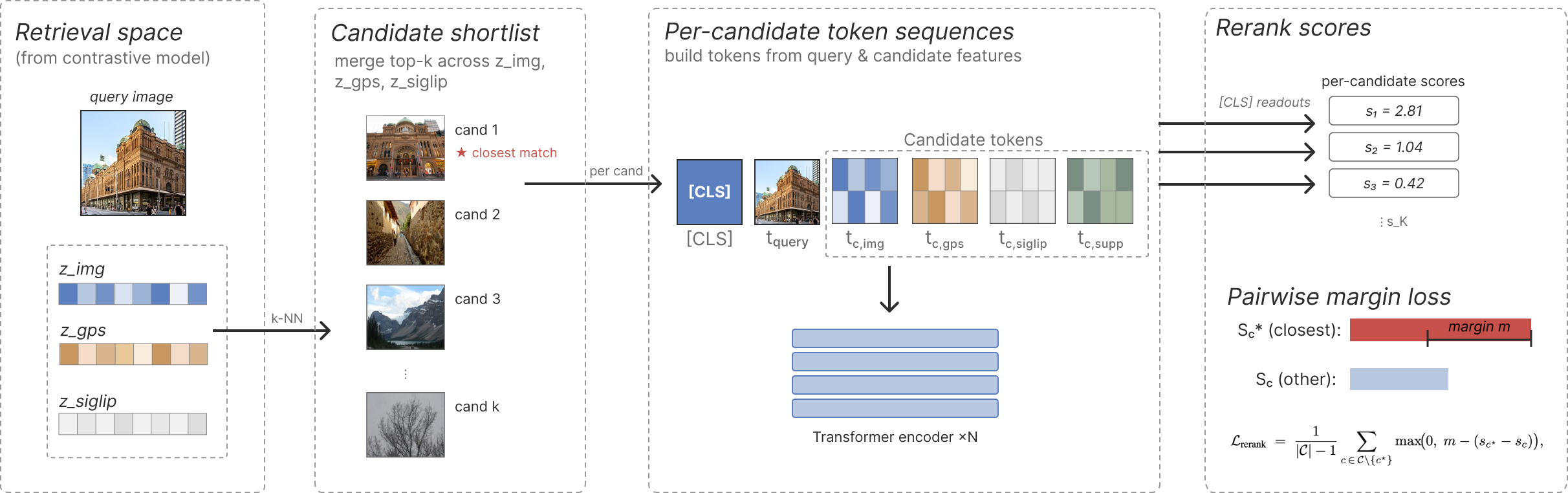}
    \caption{Attention-based reranker. Candidates from three retrieval channels are merged and de-duplicated. For each candidate, the reranker builds a short token sequence from the query representation, candidate embeddings, and local support token, then uses the final [CLS] representation to predict a rerank score. Modality-aligned retrieval metadata is included in each candidate token.}
    \label{fig:reranker}
\end{figure*}

Pure embedding-space retrieval can confuse locations that look similar at a coarse level but differ in the fine details needed for precise geolocation. To address this, we add an attention-based reranker that revisits each retrieved candidate together with the query and produces a refined score (Figure~\ref{fig:reranker}). The reranker scores candidates independently: for each candidate, the transformer attends over a small set of candidate-specific evidence rather than over the full shortlist.

\paragraph{Candidate shortlist construction.}
Given a query image, we retrieve candidate locations from a fixed retrieval index using three complementary channels with budgets $k_{\text{img}}$, $k_{\text{gps}}$, and $k_{\text{siglip}}$, respectively. The first channel retrieves candidate locations whose corresponding image embeddings are nearest to the query image embedding in the learned geographic image embedding space. The second retrieves candidate locations whose GPS embeddings are nearest to the same query image embedding. The third retrieves locations with the most visually similar reference images using frozen pre-projection SigLIP~2 embeddings. The first two channels emphasize geolocation-aware similarity, while the third preserves raw visual similarity from the backbone. We take the union of these retrieval results, merge duplicates, and keep the resulting candidate set $\mathcal{C}$. During training, because the query image comes from the same dataset used for retrieval, we mask out its indexed copy before selecting candidates. This simulates the inference setting, in which the query image is not present in the reference dataset.

\paragraph{Per-candidate token sequence.}
For each shortlisted candidate $c \in \mathcal{C}$, we construct the token sequence
\begin{equation}
\bigl[\,\text{CLS},\; t_{\text{query}},\; t_{c,\text{img}},\;  t_{c,\text{gps}},\;t_{c,\text{siglip}},\; t_{c,\text{supp}}\,\bigr].
\end{equation}
Here $t_{\text{query}}$ is the projected query image embedding from the contrastive image tower. The remaining tokens are candidate-specific. We use three candidate embeddings: the learned geographic image embedding $z_{c,\text{img}}$, the GPS embedding $z_{c,\text{gps}}$, and the frozen SigLIP visual embedding $z_{c,\text{siglip}}$,. For each modality we also compute three scalar retrieval features: the retrieval similarity $\sigma(\cdot)$, the normalized rank feature $\rho(\cdot)$, and a binary indicator $m(\cdot)$ for whether the candidate appeared in that modality's top-$k$ list. We concatenate each modality's metadata directly to its corresponding embedding before projection:
\begin{align}
t_{c,\text{img}} &= P_{\text{img}}\!\bigl([\,z_{c,\text{img}} ; \sigma_{\mathrm{img}}(c), \rho_{\mathrm{img}}(c), m_{\mathrm{img}}(c)\,]\bigr), \\
t_{c,\text{siglip}} &= P_{\text{siglip}}\!\bigl([\,z_{c,\text{siglip}} ; \sigma_{\mathrm{siglip}}(c), \rho_{\mathrm{siglip}}(c), m_{\mathrm{siglip}}(c)\,]\bigr), \\
t_{c,\text{gps}} &= P_{\text{gps}}\!\bigl([\,z_{c,\text{gps}} ; \sigma_{\mathrm{gps}}(c), \rho_{\mathrm{gps}}(c), m_{\mathrm{gps}}(c)\,]\bigr),
\end{align}
where $[\cdot ; \cdot]$ denotes concatenation and each $P(\cdot)$ is a modality-specific MLP projection to the reranker hidden dimension $H_{\text{rerank}}$. This construction allows each token to capture a specific modality's embedding content along with its retrieval cues.

\paragraph{Support token construction.}
We also add a support token that summarizes nearby evidence around each candidate anchor. The support token uses the complementary source to the query image and candidates; for the internet-photo task, the support token is constructed from street-view samples, and vice versa for the street-view task. For a candidate location $c$, we find up to $k_{\text{supp}}$ nearby images within radius $r_{\text{supp}}$, encode those support images with the cross-source image tower, and compare them to the query embedding. Let $z_q$ denote the query embedding and let $\mathcal{J}_c$ denote the valid support neighbors retrieved around candidate $c$, with embeddings $\{u_{c,j}:j\in\mathcal{J}_c\}$. We pool these support embeddings with softmax-normalized query--supp embedding similarity weights:
\begin{equation}
\alpha_{c,j} \;=\; \frac{\exp(z_q^\top u_{c,j})}{\sum_{\ell\in\mathcal{J}_c} \exp(z_q^\top u_{c,\ell})},
\qquad
\bar{u}_c \;=\; \sum_{j\in\mathcal{J}_c} \alpha_{c,j} u_{c,j}.
\end{equation}
This gives higher weight to nearby street-view images whose content best matches the query. We append four summary statistics to $\bar{u}_c$: best and mean query--supp similarity, normalized nearest-neighbor distance, and the fraction of support neighbors within $r_{\text{supp}}$. The resulting vector is projected to form $t_{c,\text{supp}}$.

Intuitively, on the internet-photo task, this token lets the reranker mimic how a human geolocation expert might geolocate an image using Google Street View. After identifying a candidate location, the reranker considers whether the surrounding street-level views are consistent with the query.

\paragraph{Reranker architecture.} The reranker is a pre-norm Transformer encoder with multi-head self-attention \cite{attentionisallyouneed} and hidden dimension $H_{\text{rerank}}$. It scores candidates independently: for each candidate $c$, the encoder attends only over that candidate's sequence $[\text{CLS}, t_{\text{query}}, t_{c,\text{img}},  t_{c,\text{gps}},t_{c,\text{siglip}}, t_{c,\text{supp}}]$. The final [CLS] representation is passed through a small MLP head to produce a scalar rerank score $s_c$. At inference time, we output the GPS coordinates of the highest-scoring candidate anchor.

\paragraph{Training objective.} We train the reranker with a pairwise margin ranking loss with margin $m$. For each query, we define the positive candidate as the shortlist candidate closest to the ground-truth location $y$, using Haversine distance: $c^\star = \arg\min_{c \in \mathcal{C}} d_H(c, y)$. We then encourage its score to exceed those of all other candidates:
\begin{equation}
\mathcal{L}_{\text{rerank}} \;=\; \frac{1}{|\mathcal{C}| - 1} \sum_{c \,\in\, \mathcal{C} \setminus \{c^\star\}} \max\!\bigl(0,\; m - (s_{c^\star} - s_c)\bigr).
\end{equation}

\section{Experiments}
\paragraph{Datasets.}
We train Pinpoint on MP16-Pro \cite{g3}, a curated subset of YFCC100M \cite{yfcc100m} containing 4.12M geotagged Flickr images, and OpenStreetView-5M \cite{osv5m}, a collection of 5.1M street-view captures distributed according to global spatial population density. We evaluate on two standard internet-image benchmarks, IM2GPS3k and YFCC4k \cite{im2gps3k_yfcc4k}, as well as the OSV-5M test set. A 1 km separation is enforced between the OSV-5M train and test sets, which prevents memorization.

\paragraph{Implementation details.}
We train Pinpoint separately for evaluation on the internet-image benchmarks (IM2GPS3k and YFCC4k) and the street-view benchmark (OSV-5M test set). In both cases, the retrieval model is essentially the same; the contrastive embedder still trains on both MP16-Pro and OSV-5M, with the only difference being the final weighting of the MP16-Pro- and OSV-5M-specific contrastive losses. During inference and reranking, the retrieval index stays limited to MP16-Pro samples in the internet-image version and OSV-5M samples in the street-view version. The reranker support token is constructed from OSV-5M samples when training on MP16-Pro, and vice versa when training on OSV-5M. Since the SigLIP~2 backbone is frozen, we precompute image embeddings for both training datasets and stream them from memory-mapped binaries during contrastive training. After the contrastive image--GPS embedder converges, its weights are frozen and a fixed retrieval index is built for reranker training. The index stores candidate coordinates together with the representations needed for multi-channel retrieval and as inputs to the reranker. All experiments were done on an NVIDIA RTX 5090 GPU.

\subsection{Comparison with State-of-the-Art}
\label{evals}
Table~\ref{tab:geoloc_results} compares Pinpoint with prior worldwide image geolocation methods on IM2GPS3k and YFCC4k, using the standard metric of the percentage of queries localized within fixed distance thresholds. Pinpoint achieves the best results across all thresholds on both benchmarks. On IM2GPS3k, it improves over the previous best method, GeoRanker, by 0.9--2.7 percentage points across thresholds, with similar gains (0.1--3.2) on YFCC4k. Among the methods compared, the three most recent methods that reported state-of-the-art results, Img2Loc, G3, and GeoRanker, all rely on prompting proprietary MLLMs. In contrast, Pinpoint does not depend on an external API and can be deployed privately, which is important for use cases involving sensitive or confidential data.

\begin{table}[t]
\centering
\caption{Comparison of our method against state-of-the-art geolocation approaches. Results are reported as the percentage of correctly localized images within the given distance thresholds.}
\label{tab:geoloc_results}
\vspace{6pt}
\setlength{\tabcolsep}{3pt} 
\small
\begin{tabular}{l *{10}{c}}
\toprule
\multirow{3}{*}{Methods} & \multicolumn{5}{c}{IM2GPS3k (\%) $\uparrow$} & \multicolumn{5}{c}{YFCC4k (\%) $\uparrow$} \\
\cmidrule(lr){2-6} \cmidrule(lr){7-11}
& Street & City & Region & Country & Continent & Street & City & Region & Country & Continent \\
& 1km & 25km & 200km & 750km & 2500km & 1km & 25km & 200km & 750km & 2500km \\
\midrule
PlaNet     \cite{planet}    & 8.5  & 24.8 & 34.3 & 48.4 & 64.6 & 5.6  & 14.3 & 22.2 & 36.4 & 55.8 \\
CPlaNet     \cite{cplanet}    & 10.2 & 26.5 & 34.6 & 48.6 & 64.6 & 7.9  & 14.8 & 21.9 & 36.4 & 55.5 \\
Translocator \cite{translocator}   & 11.8 & 31.1 & 46.7 & 58.9 & 80.1 & 8.4  & 18.6 & 27.0 & 41.1 & 60.4 \\
GeoDecoder    \cite{geodecoder}  & 12.8 & 33.5 & 45.9 & 61.0 & 76.1 & 10.3 & 24.4 & 33.9 & 50.0 & 68.7 \\
GeoCLIP     \cite{geoclip}    & 14.1& 34.5& 50.7& 69.7& 83.8& 9.6 & 19.3& 32.6& 55.0 & 74.7\\
PIGEOTTO    \cite{pigeon} & 11.3 & 36.7 & 53.8 & 72.4 & 85.3 & 10.4 & 23.7 & 40.6 & 62.2 & 77.7 \\
Img2Loc    \cite{img2loc}     & 15.3& 39.8& 53.6& 69.7 & 82.8& 19.8& 30.7& 41.4 & 58.1& 74.1\\
G3    \cite{g3}          & 16.7& 40.9& 55.6& 71.2& 84.7& 24.0& 35.9& 47.0& 64.3& 78.2\\
GLOBE \cite{globe} & 9.9 & 40.2 & 56.2 & 71.5 & 82.4 & - & - & - & - & - \\
GeoRanker   \cite{georanker}    & \underline{18.8}& \underline{45.0}& \underline{61.5}& \underline{76.3}& \underline{89.3}& \underline{32.9}& \underline{43.5}& \underline{54.3}& \underline{69.8}& \underline{82.5}\\
HierLoc \cite{hierloc} & 11.3 & 43.8  & 58.4 & 74.1  & 85.1 & 8.4 & 30.2 & 43.3 & 61.7 & 75.8 \\
\midrule
\textbf{Pinpoint}            & \textbf{20.5} & \textbf{47.4} & \textbf{63.5} & \textbf{79.0} & \textbf{90.2} & \textbf{33.0} & \textbf{44.4} & \textbf{57.5} & \textbf{71.8} & \textbf{84.5} \\
\bottomrule
\end{tabular}
\end{table}

\paragraph{Inference time comparison.}
Compared to generative MLLM-based approaches, Pinpoint is substantially faster and more practical for large-scale image geolocation. Compared to purely retrieval-based approaches like GeoCLIP, Pinpoint only adds an additional 90 ms of latency  while offering higher accuracy. For Pinpoint and GeoCLIP, latency is measured end-to-end, including image encoding, retrieval, and reranking for Pinpoint. The measured GeoRanker latency 
\begin{wraptable}{r}{0.46\linewidth}
\centering
\caption{Average inference speed comparison. Pinpoint and GeoCLIP are measured end-to-end, while GeoRanker is only measured on the final-stage reranking.}
\label{tab:inference_speed}
\vspace{6pt}
\setlength{\tabcolsep}{4pt}
\renewcommand{\arraystretch}{1.15}
\small
\begin{tabular}{l c c}
\toprule
Model & Latency / Image & Speedup \\
\midrule
GeoRanker & 288 s & 1.0$\times$ \\
Pinpoint  & 0.098 s & 3000$\times$ \\
GeoCLIP  & 0.008 s & 36000$\times$ \\
\bottomrule
\end{tabular}
\vspace{-10pt}
\end{wraptable}
covers only the final reranking stage because its full reported inference pipeline cannot be recreated without the now-deprecated GPT-4V model, which is used to generate additional location candidates before ranking \cite{georanker}. These results demonstrate the efficiency and reproducibility advantages of deterministic retrieval-based architectures over MLLM-based pipelines. All inference times are measured on IM2GPS3k. Pinpoint and GeoCLIP are evaluated with batch size of 256 and GeoRanker is evaluated with batch size of 2 to fit in GPU memory.

\subsection{Evaluation on street-view images}
\label{streetview}

Although Pinpoint achieves state-of-the-art performance on standard internet-image benchmarks, we note that these benchmarks may not fully measure generalization to unseen data. MP16-Pro, IM2GPS3k, and YFCC4k are all derived from Flickr~\cite{g3,im2gps3k_yfcc4k}, a U.S.-based photo-sharing platform, which introduces geographic and cultural biases toward English-speaking, densely populated regions. Moreover, IM2GPS3k and YFCC4k were both introduced in 2017 \cite{im2gps3k_yfcc4k}, making it likely that these benchmarks appear in the internet-scale pretraining corpora used by CLIP \cite{clip}, SigLIP \cite{siglip}, and large MLLMs used in prior work.

To evaluate generalization on a more recent and geographically diverse benchmark, we evaluate Pinpoint on the OSV-5M test split. This test split uses grid-cell sampling, which divides the map into spatial cells and samples at most one image from each cell to reduce overrepresentation of densely photographed areas~\cite{osv5m}. It also ensures that no evaluation images are within 1 km of training samples, providing a stronger measure of generalization to unseen data. As shown in Table~\ref{tab:osv5m_results}, Pinpoint achieves state-of-the-art performance on OSV-5M across both sets of reported metrics: some prior work evaluates distance-threshold accuracy, while others evaluate administrative-level accuracy by reverse-geocoding predicted coordinates and comparing the resulting country, region, and city labels to ground truth. For administrative-level accuracy, Pinpoint improves over the previous best method, HierLoc~\cite{hierloc}, by 1.9--4.1 percentage points across country, region, and city accuracy. For distance-based evaluation, it reduces mean Haversine error by 118 km, with 2.1--15.4 percentage point improvements on the distance-threshold metrics.

\begin{table*}[t]
\centering
\caption{Comparison on the OSV-5M benchmark. Top block reports the official OSV-5M metric family: GeoScore, mean Haversine error, and administrative-level accuracy based on reverse-geocoded predictions. Bottom block reports distance-threshold accuracy, the same metric used in Table~\ref{tab:geoloc_results}.}
\label{tab:osv5m_results}
\vspace{6pt}
\setlength{\tabcolsep}{6pt}
\renewcommand{\arraystretch}{0.9}
\small
\begin{tabular}{l c c *{3}{c}}
\toprule
\multirow{2}{*}{Method}
& \multirow{2}{*}{GeoScore $\uparrow$}
& \multirow{2}{*}{\shortstack{Mean Error\\(km) $\downarrow$}}
& \multicolumn{3}{c}{Classification Accuracy (\%) $\uparrow$} \\
\cmidrule(lr){4-6}
& & & City & Region & Country \\
\midrule
OSV-5M Baseline~\cite{osv5m}
& 3361 & 1814 & 5.9 & 39.4 & 68.0 \\
RFM $S^2$~\cite{aroundtheworld}
& 3767 & 1069 & 5.4 & 44.2 & 76.2 \\
HierLoc (DINOv3)~\cite{hierloc,dinov3}
& 3963 & 861 & \underline{23.3} & 55.0 & 82.9 \\
\midrule
\textbf{Pinpoint (full model)}
& \textbf{4114} & \textbf{743} & \textbf{26.0} & \textbf{59.1} & \textbf{84.8} \\
Pinpoint (retrieval only)
& \underline{4035} & \underline{784} & 23.2 & \underline{57.0} & \underline{83.8} \\
\midrule[\heavyrulewidth]
\multirow{2}{*}{Method}
& \multirow{2}{*}{\shortstack{Mean Error\\(km) $\downarrow$}}
& \multicolumn{4}{c}{Distance-Threshold Accuracy (\%) $\uparrow$} \\
\cmidrule(lr){3-6}
& & 25 km & 200 km & 750 km & 2500 km \\
\midrule
LocDiff~\cite{locdiff}
& -- & 11.0 & 46.3 & 77.0 & 88.2 \\
GRE~\cite{gre}
& 1192 & 9.7 & 35.6 & 72.5 & 91.1 \\
Semivariogram GeoCLIP~\cite{semivariogramgeoclip}
& -- & 21.5 & 52.1 & 72.1 & -- \\
\midrule
\textbf{Pinpoint (full model)}
& \textbf{743} & \textbf{35.6} & \textbf{67.5} & \textbf{83.7} & \textbf{93.2} \\
Pinpoint (retrieval only)
& \underline{784} & \underline{32.1} & \underline{65.6} & \underline{82.8} & \underline{92.8} \\
\bottomrule
\end{tabular}
\end{table*}

\subsection{Ablation Study}
To isolate the contribution of each component in Pinpoint, we report ablations on YFCC4k in Table~\ref{tab:ablation_results}. Removing the OSV-5M support token from the reranker input consistently reduces accuracy, with drops of 0.7--1.7 percentage points across distance thresholds. This aligns with our intuition that local street-view context helps in determining whether a candidate location is visually consistent with the query image.

We next evaluate the importance of the reranking stage by using only the contrastive retrieval embeddings and selecting the database location with the highest image--GPS embedding similarity. This retrieval-only variant performs substantially worse across all metrics, with the largest degradation at fine-grained thresholds: at 1 km, reranking improves accuracy by 23.9 percentage points. These results indicate that reranking is especially important for resolving visually similar nearby locations that are difficult to separate in the retrieval embedding space alone. We also ablate OSV-5M supervision during contrastive training by training the retrieval model only on MP16-Pro. Removing OSV-5M images reduces retrieval accuracy by 0.3--2.0 percentage points at four of the five distance thresholds, demonstrating that street-view supervision improves the quality of the learned embeddings for MP16-Pro retrieval.

Finally, we replace the SigLIP~2-g-opt backbone with CLIP ViT-L/14~\cite{clip} for a fairer comparison with earlier MLLM-free CLIP-based methods such as PIGEON \cite{pigeon} and GeoCLIP \cite{geoclip}. At each distance threshold, we compare against the stronger result reported by either method. Even with the older CLIP backbone, Pinpoint improves accuracy by 18.0, 15.2, 9.6, 4.1, and 2.1 percentage points across the 1 km, 25 km, 200 km, 750 km, and 2500 km thresholds, respectively. These results indicate that Pinpoint’s gains are driven by the overall retrieval and reranking pipeline, rather than by the stronger pretrained backbone alone. Compared to MLLM-based approaches, this CLIP-based variant remains competitive, performing between G3 and GeoRanker.
\begin{table}[h!]
\centering
\caption{Ablation study on YFCC4k comparing Pinpoint with variants that remove support, reranking, OSV-5M supervision, or the SigLIP~2 backbone.}
\label{tab:ablation_results}
\vspace{6pt}
\setlength{\tabcolsep}{5pt}
\scriptsize
\begin{tabular}{l c *{5}{c}}
\toprule
\multirow{2}{*}{Ablation} 
& \multirow{2}{*}{\shortstack{Median Error\\(km) $\downarrow$}}
& \multicolumn{5}{c}{YFCC4k Accuracy (\%) $\uparrow$} \\
\cmidrule(lr){3-7}
& & 1 km & 25 km & 200 km & 750 km & 2500 km \\
\midrule
\textbf{Pinpoint (full model)}
& \textbf{84.9} & \textbf{33.0} & \textbf{44.4} & \textbf{57.5} & \textbf{71.8} & \textbf{84.5} \\
Pinpoint without OSV-5M support token
& 95.6 & 31.8 & 43.3 & 55.8 & 71.1 & 83.7 \\
Retrieval only (without reranker)
& 262.9 & 9.1 & 29.0 & 46.9 & 65.8 & 80.8 \\
Retrieval without OSV-5M images
& 287.4 & 9.7 & 28.4 & 44.9 & 65.1 & 80.5 \\
\midrule
Pinpoint using CLIP ViT-L/14 backbone
& 195.3 & 28.4 & 38.9 & 50.2 & 66.3 & 79.8 \\
Retrieval using CLIP ViT-L/14 backbone
& 497.8 & 7.7 & 22.3 & 36.3 & 57.0 & 74.5 \\
\bottomrule
\end{tabular}
\end{table}

We further examine the contribution of each candidate token on IM2GPS3k in Table~\ref{tab:token_ablation_results} by separately retraining the reranker with the corresponding token and its projection removed during both training and evaluation. Each variant is evaluated at 50{,}000 training steps. Removing any candidate token increases median error from 40.4 km to 47.3--50.5 km. Removing the GPS token produces the largest increase in median error, while removing the support token causes the largest decrease in 1 km accuracy, from 20.5\% to 18.2\%. These results support the complementary roles of geographic, visual, and nearby cross-source evidence in candidate selection. 

\begin{table}[h!]
\centering
\caption{Token ablation study on IM2GPS3k comparing Pinpoint with separately retrained variants that remove the candidate image, GPS, SigLIP, or support token.}
\label{tab:token_ablation_results}
\vspace{6pt}
\setlength{\tabcolsep}{5pt}
\scriptsize
\begin{tabular}{l c *{5}{c}}
\toprule
\multirow{2}{*}{Ablation}
& \multirow{2}{*}{\shortstack{Median Error\\(km) $\downarrow$}}
& \multicolumn{5}{c}{IM2GPS3k Accuracy (\%) $\uparrow$} \\
\cmidrule(lr){3-7}
& & 1 km & 25 km & 200 km & 750 km & 2500 km \\
\midrule
\textbf{Pinpoint (full model)}
& \textbf{40.4} & \textbf{20.5} & \textbf{47.4} & \textbf{63.5} & \textbf{79.0} & \textbf{90.2} \\
Pinpoint without candidate image token
& 47.3 & 20.0 & 46.3 & 62.2 & 78.2 & 89.7 \\
Pinpoint without candidate GPS token
& 50.5 & 20.4 & 46.5 & 61.6 & 78.1 & 90.0 \\
Pinpoint without candidate SigLIP token
& 49.2 & 19.8 & 46.5 & 62.3 & 78.5 & 90.2 \\
Pinpoint without support token
& 49.2 & 18.2 & 46.6 & 62.6 & 78.3 & 90.1 \\
\bottomrule
\end{tabular}
\end{table}

\subsection{Hyperparameter Analysis}
We sweep several key hyperparameters in the Pinpoint pipeline to better understand their impact. Specifically, we vary the cross-source loss weight $\lambda$ and the number of retrieval candidates selected from each of the three shortlist channels: learned image, GPS, and SigLIP retrieval. Unless otherwise specified, we use our model's default configuration: $\lambda=1/3$, $k_{\text{img}}=8$, $k_{\text{siglip}}=2$, and $k_{\text{gps}}=2$. To avoid tuning on the test benchmarks used in our main evaluation, we conduct these sweeps on a separate benchmark, IM2GPS \cite{im2gps}.

\begin{figure*}[t]
\centering
\begin{tabular}{@{}c@{\hspace{0.02\textwidth}}c@{}}
\includegraphics[height=0.12\textheight]{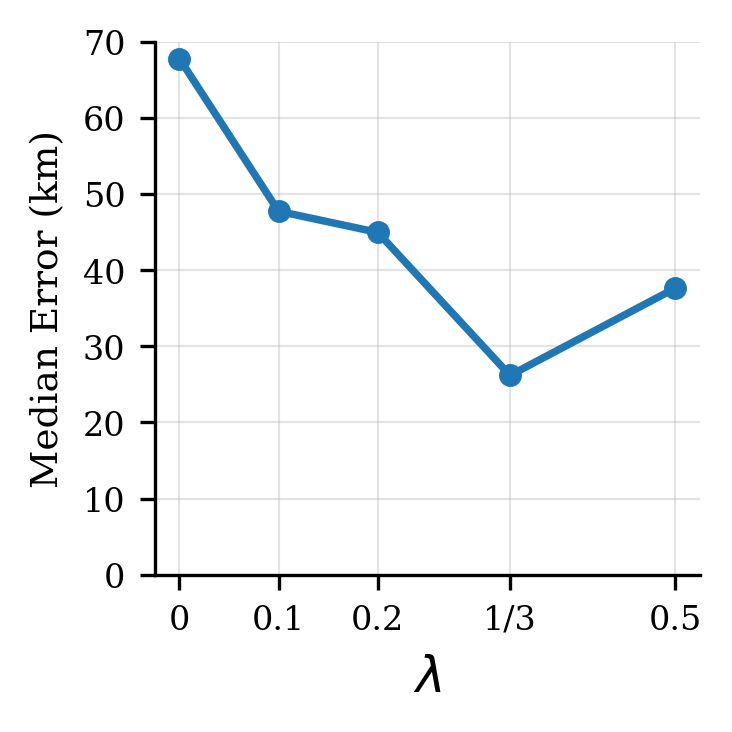} &
\includegraphics[height=0.12\textheight]{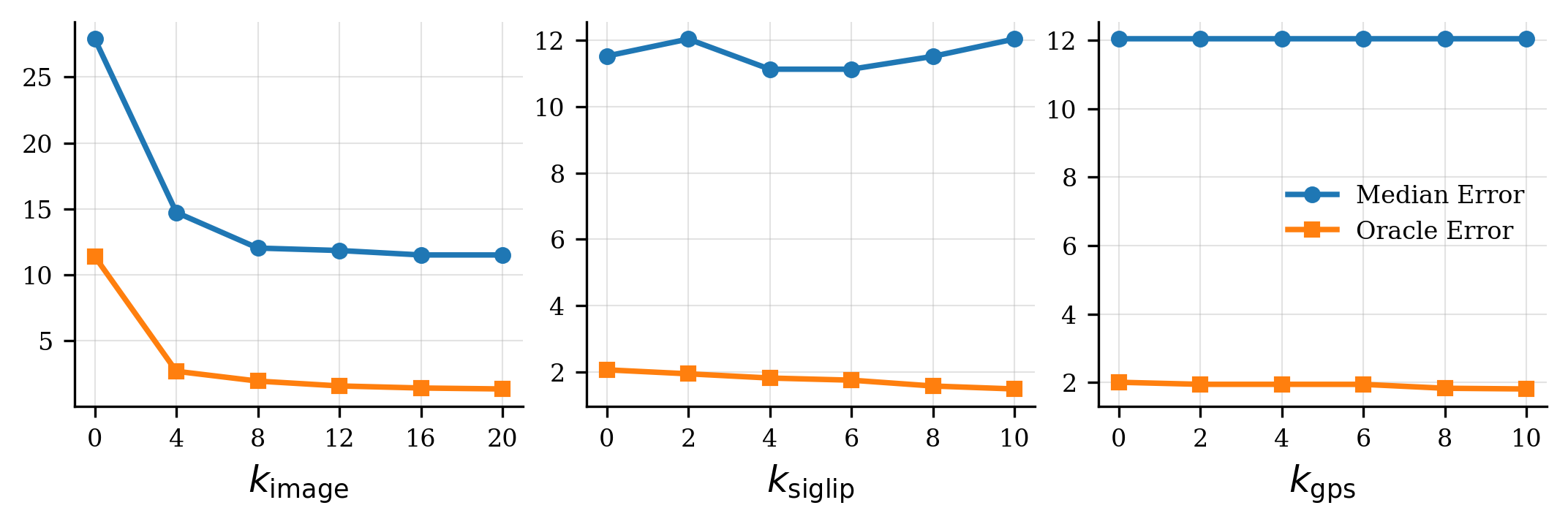}
\end{tabular}
\caption{Hyperparameter sweeps on IM2GPS. Left: median error of retrieval-only model vs OSV-5M mixing weight $\lambda$. Right: median and oracle error under different reranker candidate budgets. Candidate-budget sweeps vary one retrieval channel at a time while keeping the other channels fixed.}
\label{fig:hyperparameter_sweeps}
\end{figure*}

Figure~\ref{fig:hyperparameter_sweeps} shows that adding OSV-5M supervision substantially reduces median error, with the best performance at $\lambda=1/3$. Median error increases when the two sources are weighted equally at $\lambda=0.5$, suggesting that cross-source training is beneficial up to a point, after which over-weighting OSV-5M can hurt target-domain performance. We also study the candidate budgets used to form the reranker shortlist, varying one channel at a time while keeping the others fixed. Increasing the number of candidates in each channel monotonically reduces oracle error, but does not always reduce median error after reranking. This is because larger shortlists improve candidate recall while making selection more difficult. We observe diminishing returns as candidate budgets increase, especially for $k_{\text{siglip}}$ and $k_{\text{gps}}$, where median error changes by less than 1 km across the sweep. Overall, our default setting provides a good trade-off between candidate recall and reranking accuracy.

\section{Conclusion}
We presented Pinpoint, a retrieve-and-rerank architecture for worldwide image geolocation that incorporates both ordinary internet photos and street-view imagery within a single coarse-to-fine pipeline. Pinpoint achieves state-of-the-art results across all distance thresholds on IM2GPS3k, YFCC4k, and the OSV-5M test set. We showed that cross-source training and reranking each contribute substantially to final performance, and that strong worldwide geolocation does not require prompting a multimodal large-language model at inference time. 

\textbf{Limitations.}
Although Pinpoint is low-latency at inference, its frozen SigLIP 2 backbone requires a one-time precomputation of image embeddings for every training and reference entry, which must be stored persistently. The retrieval index also introduces two structural limitations: predictions are constrained to indexed locations, and the index is temporally static, which means visual changes such as new construction are not reflected until it is rebuilt. The coverage limitation can be mitigated with a large, comprehensive index, and temporal accuracy can be maintained by periodically updating it with new images.

\textbf{Broader Impact.}
Image geolocation has clear beneficial applications, including disaster response, navigation, biodiversity and ecological monitoring, and open-source investigative work. However, there are also potential ethical concerns inherent to image geolocation, such as misuse for surveillance, stalking, or doxxing. For this reason, we release our code for academic analysis, but chose not to publicly release trained model weights.

\begin{ack}
The authors received no third-party funding for this work and declare no competing interests.
\end{ack}

\newpage

\bibliographystyle{plainnat}
\bibliography{references}

\newpage
\appendix
\section{Implementation Details}
\label{app:implementation_details}

Table~\ref{tab:training} reports the optimization settings used in the main experiments. The remaining implementation parameters are grouped by component below.

\begin{table}[h!]
    \centering
    \caption{Optimization hyperparameters for the contrastive embedder and attention reranker. Both stages use linear warmup to peak learning rate followed by cosine decay over the remaining steps. Note that we train the street view model for double the number of steps; 100k instead of 50k.}
    \label{tab:training}
    \vspace{6pt}
    \small
    \begin{tabular}{lrr}
        \toprule
                                 & \textbf{Embedder} & \textbf{Reranker} \\
        \midrule
        Steps                    & 50{,}000          & 50{,}000          \\
        Warmup steps             & 1{,}000           & 1{,}000           \\
        Batch size               & 4{,}096           & 512               \\
        Peak learning rate       & $1\mathrm{e}{-3}$ & $1\mathrm{e}{-4}$ \\
        Weight decay             & $1\mathrm{e}{-4}$ & $1\mathrm{e}{-4}$ \\
        LR schedule              & Cosine            & Cosine            \\
        Optimizer~\cite{adamw}             & AdamW             & AdamW             \\
        Mixed precision          & bf16              & bf16              \\
        \bottomrule
    \end{tabular}
\end{table}

\paragraph{Backbone and embedder.}
The frozen visual encoder is SigLIP~2 giant-opt patch16 384. The shared embedding dimension is $d=256$. Image adapters use 2 residual blocks, hidden width 1024, dropout 0.15, and expansion factor 2. The shared image trunk uses 4 residual blocks, hidden width 1024, dropout 0.15, and expansion factor 2. The location tower uses 6 residual blocks per scale, hidden width 1024, dropout 0.15, and expansion factor 2. The learned Fourier feature bank uses $S=3$ scales with 128 features per scale and initialization scales 1.0, 16.0, and 256.0. In the main text, $\lambda$ denotes the OSV-5M mixing weight in the contrastive loss; the value is $\lambda=1/3$ for the internet-photo model and $\lambda=0.8$ for the street-view model.

\paragraph{Candidate retrieval and support.}
The reranker shortlist uses $k_{\text{img}}=8$, $k_{\text{siglip}}=2$, and $k_{\text{gps}}=2$, with a candidate-pool multiplier of 2 and at most 12 merged candidates. The retrieval index is built from the training split with batch size 65{,}536 and bfloat16 embeddings. For each candidate, the support token uses up to $k_{\text{supp}}=4$ nearby images within radius $r_{\text{supp}}=1.0$ km. Its summary statistics are best query--supp similarity, mean query--supp similarity, normalized nearest-neighbor distance, and support fraction.

\paragraph{Reranker.}
The reranker Transformer encoder has 4 layers, 8 attention heads, hidden dimension $H_{\text{rerank}}=512$, feed-forward dimension 2048, and dropout 0.2. Reranker sequence chunk size is 16{,}384. The ranking objective is a pairwise margin loss with margin $m=0.2$.

\section{Retrieval Supervision}
\label{app:retrieval_supervision}
We compare image--GPS supervision with image--image SigLIP variants that select either the nearest or a random same-source image within 1 km as the positive (Table~\ref{tab:retrieval_supervision}).

\begin{table}[h!]
\centering
\caption{Retrieval diagnostics on IM2GPS3k under image--image and image--GPS supervision. Distance-threshold values are accuracy percentages.}
\label{tab:retrieval_supervision}
\vspace{6pt}
\setlength{\tabcolsep}{5pt}
\scriptsize
\begin{tabular}{l l c *{5}{c}}
\toprule
\multirow{2}{*}{Supervision}
& \multirow{2}{*}{Retrieval score}
& \multirow{2}{*}{\shortstack{Median Error\\(km) $\downarrow$}}
& \multicolumn{5}{c}{IM2GPS3k Accuracy (\%) $\uparrow$} \\
\cmidrule(lr){4-8}
& & & 1 km & 25 km & 200 km & 750 km & 2500 km \\
\midrule
Image--image (nearest positive) & Image
& 131.9 & 14.9 & 39.3 & 54.4 & 72.0 & 86.6 \\
Image--image (random positive <1 km) & Image
& 82.9 & 16.8 & 42.8 & 57.9 & 74.4 & 87.3 \\
Image--GPS & GPS
& 81.1 & 8.7 & 40.2 & 60.1 & 77.2 & 88.8 \\
Image--GPS & Image
& 94.4 & 15.7 & 41.5 & 57.5 & 74.3 & 87.4 \\
Image--GPS & 0.7 image + 0.3 GPS
& 54.9 & 17.2 & 45.4 & 61.6 & 77.5 & 89.2 \\
\bottomrule
\end{tabular}
\end{table}

For the image--GPS model, GPS similarity favors coarse-scale accuracy, while image similarity improves accuracy at 1 km. Combining their scores improves every reported distance-threshold accuracy over either channel alone, illustrating their complementary geographic and visual information.

\section{Reference Corpus Coverage}
\label{app:corpus_coverage}
We measure reference coverage using an oracle that selects the indexed coordinate closest to each query's ground truth (Table~\ref{tab:corpus_oracle_results}). Most internet-photo queries have a reference within 1 km, and over 98\% within 25 km. OSV-5M has almost no sub-kilometer matches due to its train/test separation. Actual performance also depends on candidate retrieval and reranking.

\begin{table}[h!]
\centering
\caption{Corpus-oracle accuracy on the three evaluation benchmarks, computed using Haversine distance over the training reference corpora.}
\label{tab:corpus_oracle_results}
\vspace{6pt}
\setlength{\tabcolsep}{6pt}
\small
\begin{tabular}{l *{5}{c}}
\toprule
\multirow{2}{*}{Benchmark}
& \multicolumn{5}{c}{Accuracy (\%) $\uparrow$} \\
\cmidrule(lr){2-6}
& 1 km & 25 km & 200 km & 750 km & 2500 km \\
\midrule
IM2GPS3k & 86.45 & 99.00 & 100.00 & 100.00 & 100.00 \\
YFCC4k & 93.67 & 99.69 & 99.98 & 99.98 & 100.00 \\
OSV-5M & 0.14 & 86.35 & 99.38 & 99.94 & 100.00 \\
\bottomrule
\end{tabular}
\end{table}

\section{Example Results}

\begin{figure}[h]
    \centering
    \includegraphics[width=0.9\linewidth]{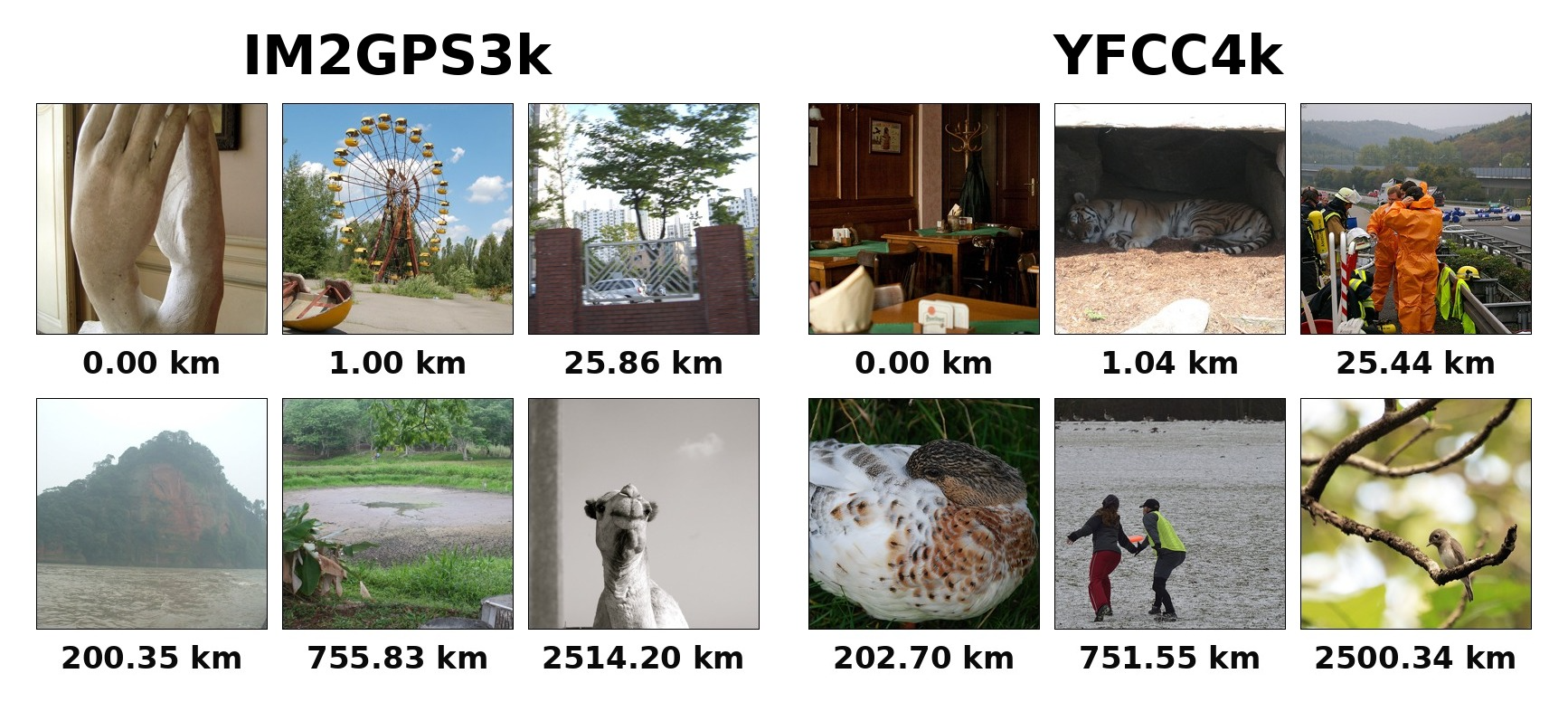}
    \caption{Pinpoint's distance errors across example images from IM2GPS3k and YFCC4k. Each image caption represents how far the prediction was from the ground truth. Examples are selected from each distance threshold used for evaluation.}
    \label{fig:internet-examples}
\end{figure}

\begin{figure}[h]
    \centering
    \includegraphics[width=0.5\linewidth]{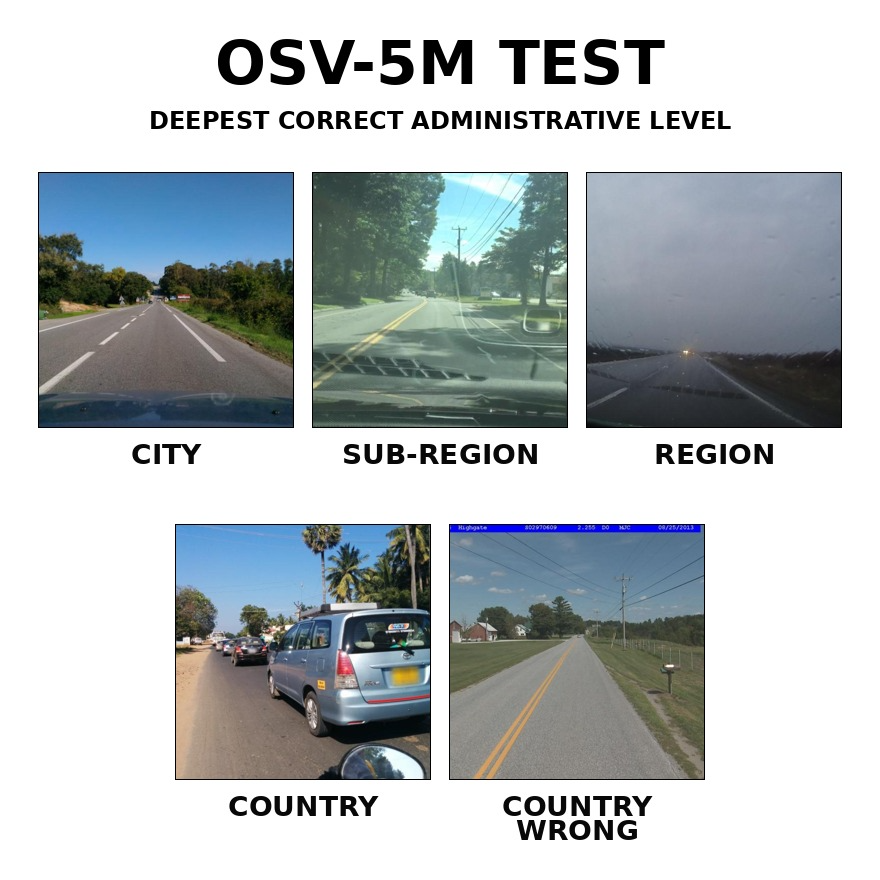}
    \caption{Example images from the OSV-5M test set for which Pinpoint correctly identified the city, sub-region, region, country, and for which Pinpoint mislabeled the country.}
    \label{fig:placeholder}
\end{figure}

\section{Existing Assets}
\label{app:data_assets}

We use MP16-Pro~\cite{g3}, YFCC100M~\cite{yfcc100m}, OSV-5M~\cite{osv5m}, SigLIP~2 giant-opt patch16 384~\cite{siglip2}, and OpenAI CLIP ViT-L/14~\cite{clip}. We do not redistribute dataset images. Users should obtain each dataset from its original source and follow the corresponding terms; MP16-Pro and other Flickr-derived datasets are governed by original Flickr media licenses or dataset terms at the item level. YFCC100M uses item-level Creative Commons licenses, OSV-5M is listed as CC-BY-SA-4.0, SigLIP~2 is listed as Apache-2.0, and OpenAI CLIP is released under the MIT license.

\end{document}